\documentclass[letterpaper, 10 pt, conference]{ieeeconf}  % Comment this line out if you need a4paper
\makeatletter
\IEEEtriggercmd{\reset@font\normalfont\fontsize{7.8pt}{8.50pt}\selectfont}
\makeatother
\IEEEtriggeratref{1} % 从第一个引用条目开始生效

\IEEEoverridecommandlockouts                              % This command is only needed if 
\usepackage{booktabs}
\usepackage{hyperref}
\usepackage{graphicx}
\usepackage{algorithm}
\usepackage{algpseudocode}
\usepackage{amsmath}
\usepackage{caption}
\usepackage{graphicx}
\usepackage{subcaption}
\usepackage[T1]{fontenc}
\usepackage{textcomp}
\usepackage{tabularx}
\usepackage{booktabs}
\usepackage{array}
\usepackage{ragged2e}

\title{\LARGE \bf
ExploreVLM: Closed-Loop Robot Exploration Task Planning with Vision-Language Models
}

\author{Zhichen Lou, Kechun Xu, Zhongxiang Zhou and Rong Xiong% <-this % stops a space
\thanks{Rong Xiong, Zhichen Lou and Kechun Xu are with the State Key Laboratory of Industrial Control Technology, and Institute of Cyber-Systems and Control, Zhejiang University, Hangzhou 310027, China.}% <-this % stops a space
\thanks{Zhongxiang Zhou is with Zhejiang Humanoid Robot Innovation Center Co., Ltd.}
}

\begin{document}

\maketitle
\thispagestyle{empty}
\pagestyle{empty}

%%%%%%%%%%%%%%%%%%%%%%%%%%%%%%%%%%%%%%%%%%%%%%%%%%%%%%%%%%%%%%%%%%%%%%%%%%%%%%%%
\begin{abstract}

The advancement of embodied intelligence is accelerating the integration of robots into daily life as human assistants. This evolution requires robots to not only interpret high-level instructions and plan tasks but also perceive and adapt within dynamic environments. Vision-Language Models (VLMs) present a promising solution by combining visual understanding and language reasoning. However, existing VLM-based methods struggle with interactive exploration, accurate perception, and real-time plan adaptation. To address these challenges, we propose ExploreVLM, a novel closed-loop task planning framework powered by Vision-Language Models (VLMs). The framework is built around a step-wise feedback mechanism that enables real-time plan adjustment and supports interactive exploration. At its core is a dual-stage task planner with self-reflection, enhanced by an object-centric spatial relation graph that provides structured, language-grounded scene representations to guide perception and planning. An execution validator supports the closed loop by verifying each action and triggering re-planning. Extensive real-world experiments demonstrate that ExploreVLM significantly outperforms state-of-the-art baselines, particularly in exploration-centric tasks. Ablation studies further validate the critical role of the reflective planner and structured perception in achieving robust and efficient task execution.

\end{abstract}

%%%%%%%%%%%%%%%%%%%%%%%%%%%%%%%%%%%%%%%%%%%%%%%%%%%%%%%%%%%%%%%%%%%%%%%%%%%%%%%%

\section{INTRODUCTION}

The evolution of embodied intelligence is ushering robots into everyday scenarios as assistants for human activities. This requires systems that can interpret high-level instructions, perceive dynamic environments, and adapt plans online to achieve task objectives effectively. 

Vision-Language Models (VLMs), which integrate visual understanding and linguistic reasoning, offer a promising direction for robot task planning \cite{zeng2023_survey}. They enable extraction of task-relevant information from images and grounding of structured descriptions in perception \cite{saycan,voxposer,innermonologue,digknow,seedo}. Recent efforts such as VILA \cite{VILA} for scene-aware planning and ReplanVLM \cite{ReplanVLM} for planning with error correction have leveraged VLMs for planning and perception.

\begin{figure}[htbp]
  \centering
  \includegraphics[width=\linewidth]{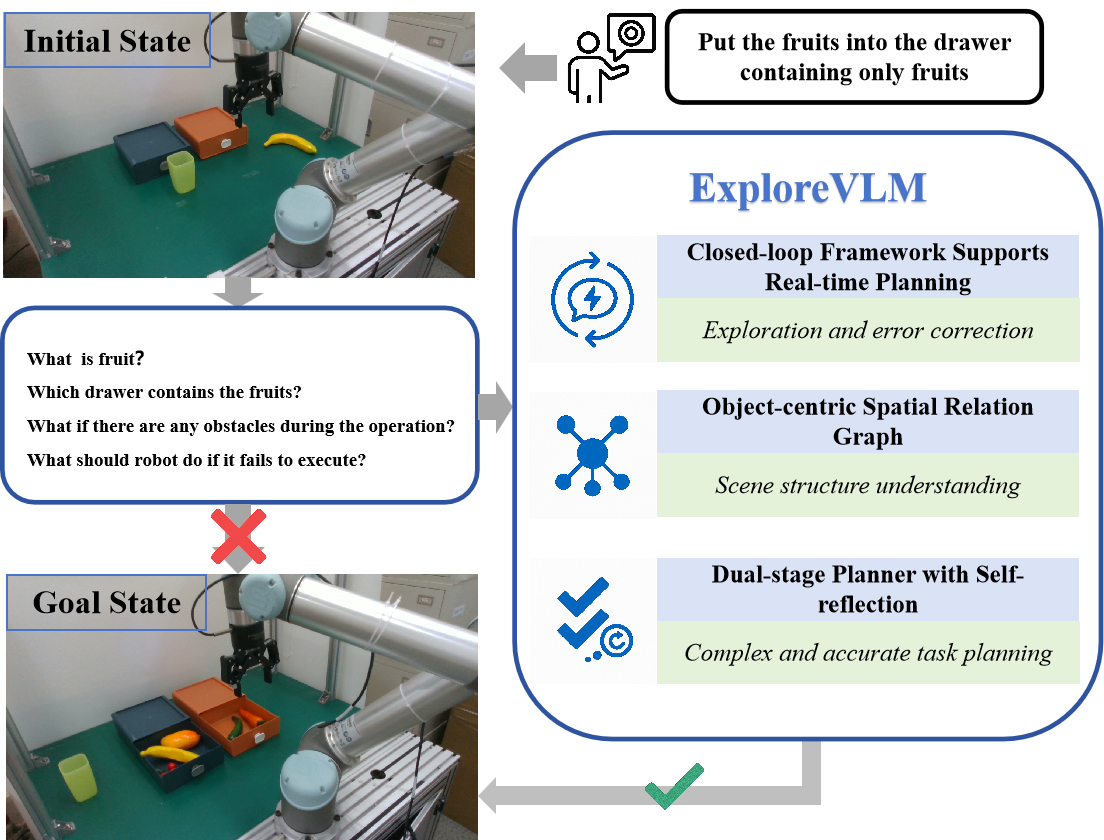} 
  \caption{Given the task goal, traditional task-planning frameworks struggle with identifying fruits, locating the correct drawer, handling obstacles, and managing failures. \textbf{ExploreVLM} overcomes these challenges via closed-loop real-time planning, object-centric spatial relation graph, and a dual-stage planner with self-reflection.}
  \label{fig:introduction}
\vspace{-1em}
\end{figure}

However, existing VLM-based approaches often struggle with real-world, interactive tasks that require closed-loop perception, adaptive planning, and multi-step reasoning \cite{RoboExp}. Consider the task in fig \ref{fig:introduction}: "Put the fruits into the drawer containing only fruit." To solve this, the robot must first identify which objects in the environment are fruits through visual and semantic understanding. Next, it must explore the scene to locate and open drawers, then inspect their contents to determine which drawer is exclusively for fruit. Throughout this process, the robot may encounter occlusions or obstacles blocking access to the drawers—requiring it to reason about and remove interfering objects before completing the task. Each interaction may alter the scene, demanding real-time perception updates and dynamic plan adjustments. Current VLM pipelines, which typically operate in an open-loop manner with static perceptions and single-shot planning, often fail to robustly solve such complex, exploration-centric tasks in partially observed environments.

\begin{figure*}[!t]
  \centering
  \includegraphics[width=\linewidth]{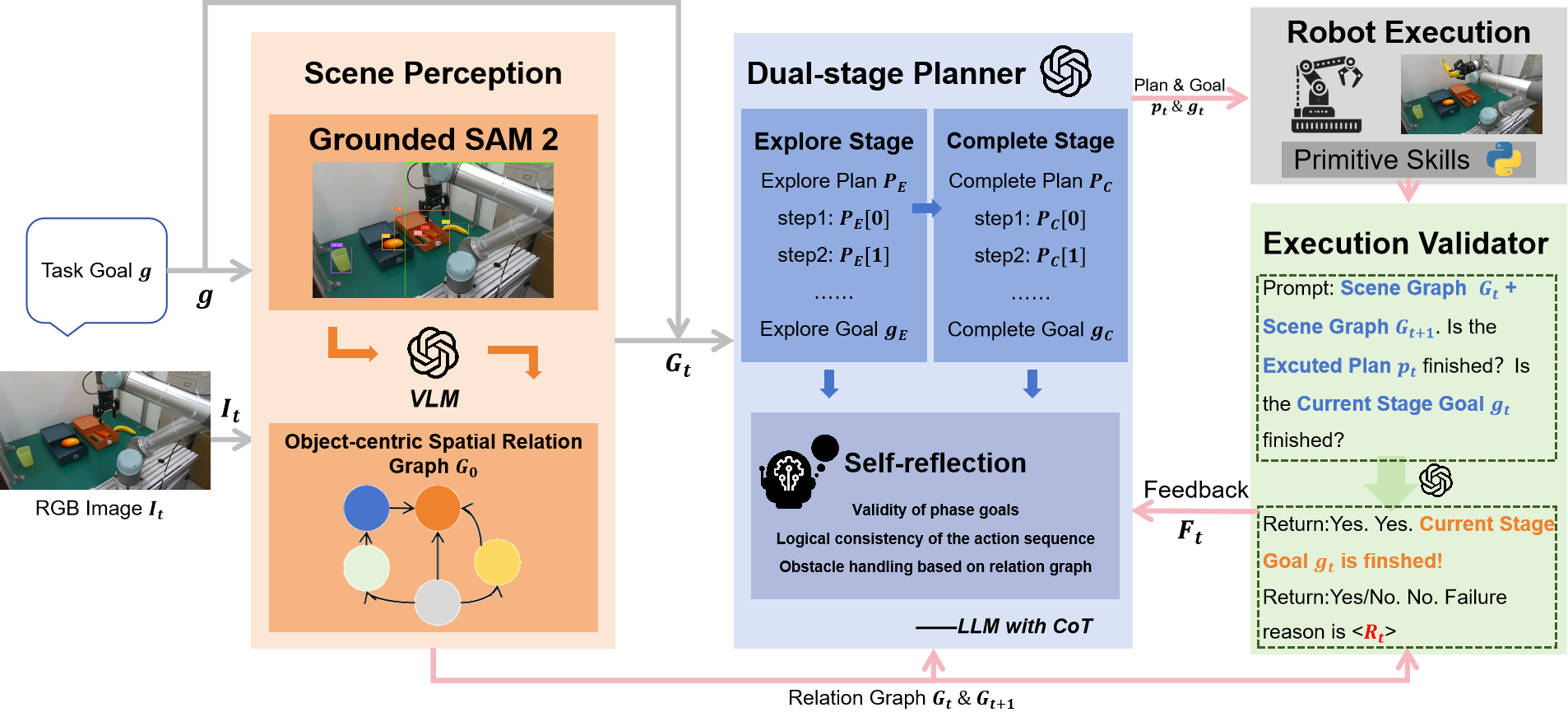} % 替换为你的文件名（不带后缀）
  \caption{Overview of the \textbf{ExploreVLM} framework. Given an image $I_t$ and task goal $g$, the scene perception module extracts an object-centric spatial relation graph $G_t$. Then the dual-stage planner generates sub-goals and action plans in explore-complete stage with self-reflection. After robot executed the plan $p_t$, the validator sends feedback $F_t$ to the planner. The pink loop indicates the closed-loop process, where perception, planner, execution, and validator work together to finish the task.}
  \label{fig:framework}
\vspace{-1em}
\end{figure*}

To address these challenges, we propose ExploreVLM, a novel closed-loop task planning framework powered by Vision-Language Models. ExploreVLM unifies structured scene perception, high-level planning, and per-step execution validation, enabling robots to perform real-time, feedback-driven adaptation. The framework features a dual-stage planner: an exploration stage actively seeks missing task-relevant information (e.g., finding coke, opening and inspecting drawers), while a completion stage focuses on achieving the overall goal. An object-centric spatial relation graph provides structured, language-grounded scene representations, making object categories, relations, and action preconditions explicit for the planner. Additionally, a self-reflection mechanism reviews and refines plans to avoid infeasible or illogical steps, while an execution validator checks the outcome of each action and triggers re-planning—enabling robust and interactive exploration. Our main contributions are summarized as follows:

1. We propose \textbf{ExploreVLM, a Vision-Language Model-based closed-loop task planning framework for robots}. It integrates scene perception, high-level task planning, and execution validation into a unified system. ExploreVLM leverages the visual and linguistic reasoning capabilities of VLMs to extract structured task-relevant information from RGB images and high-level instructions, and uses per-step feedback to enable real-time plan adaptation. This closed-loop design supports flexible, robust decision-making in dynamic environments.

2. We design a \textbf{dual-stage task planner with self-reflection}, enhanced by an \textbf{object-centric spatial relation graph} for structured scene perception. The graph models spatial relationships among objects to ground visual scenes in text, while the planner decomposes tasks into "explore-complete" stages and refines plan logic through self-reflection, improving accuracy under uncertainty.

3. We \textbf{validate ExploreVLM on a real-world robotic platform}. Experiments demonstrate superior performance over strong baselines, particularly in exploration tasks. Ablation studies confirm the critical roles of the object-centric spatial relation graph and the dual-stage planner with self-reflection in boosting task success rate and execution robustness.

\section{Related Work}

Traditional robot task planning has been extensively studied using Task and Motion Planning (TAMP) frameworks. Kaelbling and Lozano-Pérez \cite{kaelbling2011} and Garrett et al. \cite{garrett2021} introduced hierarchical and integrated approaches to combine low-level continuous motion planning with high-level discrete task planning, enabling robots to execute long-horizon tasks efficiently. However, these classical TAMP methods typically emphasize symbolic planning or optimization-based strategies, such as those proposed by Toussaint \cite{toussaint2015} and Toussaint et al. \cite{toussaint2018}, without directly integrating natural language instructions or visual environmental perception. For example,, while Srivastava et al. \cite{srivastava2022} explored household task behavior using BDDL. Despite these advances, such methods are limited in adaptability, especially in dynamic or open-world environments.

With the rapid advancement of Vision-Language Models (VLMs) \cite{openai2023,hurst2024,wang2024}, recent research has begun to explore their application in robot task planning. VLMs’ ability to jointly understand images and language enables more flexible and generalizable planning solutions. Silver et al. \cite{silver2024} leveraged pretrained large language models (LLMs) to synthesize Python programs and generate task plans in the planning domain definition language (PDDL). Ahn et al. \cite{saycan} introduced SayCan, which integrates LLMs with executable pre-trained skills by leveraging value functions and context awareness for closed-loop language-action control. Huang et al. \cite{voxposer} proposed VoxPoser, a framework that extracts affordances and constraints for manipulation tasks, employing LLMs to interact with VLMs and construct composable 3D value maps, thereby supporting zero-shot trajectory generation. In addition, Huang et al. \cite{innermonologue} developed Inner Monologue, using language feedback (e.g., success detection, scene description) as an "inner monologue" to improve LLM reasoning and adaptability, significantly boosting task success rates.
The use of human demonstration videos as sources of generalizable knowledge, rather than explicit instructions, has also attracted attention. Chen et al. \cite{digknow} proposed the DigKnow framework to extract task knowledge from observation, action, and pattern levels, which is then combined with LLMs for robust task planning and execution verification. Similarly, Wang et al. \cite{seedo} presented SeeDo, converting human demonstration videos into executable plans by integrating vision-language understanding at multiple knowledge levels.

Several comprehensive VLM-based frameworks have sought to overcome these limitations. Hu et al. \cite{VILA} developed VILA, which utilizes GPT-4V to directly integrate perceived visual data and textual instructions for plan generation. Mei et al. \cite{ReplanVLM} introduced ReplanVLM, which focuses on error correction and re-planning through both internal and external mechanisms during task execution. 
However, while these approaches support advanced task planning, they still lack precise scene perception and robust interactive exploration capabilities, especially for complex open-ended tasks. To address the need for interactive scene understanding, Jiang et al. \cite{RoboExp} proposed RoboExp, which employs multimodal models and explicit memory to construct Action-Conditioned Scene Graphs (ACSGs) for manipulation. Nevertheless, RoboExp is constrained by its reliance on prior object knowledge and extensive exploratory steps, often resulting in unnecessary actions.

To overcome the above challenges, we propose ExploreVLM, a novel framework that fully leverages VLMs for real-time, closed-loop integration of scene perception, dual-stage task planning with self-reflection, and execution validation. Our method introduces an object-centric spatial relation graph for scene representation and a dual-stage planning module specifically designed to address uncertainty, diversity, and complexity in unstructured environments.

\section{Method}
\subsection{Overall Framework of ExploreVLM} 
As illustrated in Fig. \ref{fig:framework}, the ExploreVLM framework takes an initial RGB image $I_0$ and a language task goal $g$ (e.g., "put the fruits into the drawer containing only fruits") as inputs. First, the image and goal are fed into the scene perception module, which extracts the current environmental information into an object-centric spatial relation graph $G_\mathrm{0}$. This graph explicitly captures the relative spatial positions between all detected objects. Next, the graph $G_\mathrm{0}$ and the goal $g$ are passed to the dual-stage task planner with self-reflection. This planner decomposes the overall task into two sequential stages: an exploration stage (if necessary to locate objects or gather missing information) and a completion stage. For each stage, the module defines specific sub-goals $g_E$, $g_C$ and generates a corresponding sequence of low-level robot actions $P_E$, $P_C$. The initial plan is then optimized through a self-reflection mechanism to enhance its logical consistency and feasibility. Subsequently, the generated plan is sent to the robot execution module. If an exploration stage exists, the robot executes its steps first, aiming to confirm the location or state of target objects.

\begin{algorithm}[htbp]
\caption{ExploreVLM}
\label{alg:algorithm1}
\begin{algorithmic}[1]
\Require Initial RGB image $I_0$, language task goal $g$ and primitive skills $\Pi$
\State $t = 0$, $F_\mathrm{0} = None$
\While{True}
    \State $G_t = \text{ScenePerception}(I_t, g)$ \Comment{Extract object-centric relation graph}
    \State $(P_E, g_E, P_C, g_C) = \text{TaskPlanner}(G_t, g, F_t)$ \Comment{Generate dual-stage plans and goals}
    \If{$P_E = g_E = P_C = g_C = None$}
        \State \textbf{break} \Comment{Task finish}
    \EndIf
    \If{$P_E \neq \text{None}$ and $g_E \neq \text{None}$}
        \State $(p_t, g_t) = (P_E[0], g_E)$ \Comment{Need to explore first}
    \Else
        \State $(p_t, g_t) = (P_C[0], g_C)$ \Comment{No need to explore}
    \EndIf
    \State Execute skill $\Pi(p_t)$ \Comment{Execute current plan}
    \State $G_{t+1} = \text{ScenePerception}(I_{t+1}, g)$ \Comment{Update relation graph}
    \State $F_t = \text{ExecutionValidator}(G_t, G_\mathrm{t+1}, _t, p_t, g_t)$ \Comment{Feedback}
    \State $t = t + 1$
\EndWhile
\end{algorithmic}
\end{algorithm}

Critically, mirroring human task execution, after each action execution, the execution validator will check whether the current sub-goal $g_t$ associated with that step $p_t$ has been achieved. Then it provides step-level feedback $F_t$ to the dual-stage task planner. Based on this feedback $F_t$, the planner dynamically adjusts the plan—enabling re-execution, revision, or progression as needed. Once the exploration stage yields sufficient information, the planner updates the subsequent completion stage accordingly. The system proceeds in a closed loop until the task is successfully completed. This closed-loop orchestration of the three modules through per-step feedback after action execution, enables robust adaptation and ultimately drives the system towards successful task completion. 

The complete process is formalized in Algorithm \ref{alg:algorithm1}, where $t$ is the time step, $F_t$ is the feedback from the validator at time $t$, $G_t$ is the object-centric spatial relation graph extracted by the scene perception module at time $t$, $I_t$ is the RGB image at time $t$, $g$ is the input task goal, $P_E$, $g_E$, $P_C$, $g_C$ is the stage plan sequences and sub-goals of exploration and complete stage generated by dual-stage planner with reflection, $p_t$, $g_t$ is the plan and associated sub-goal to be executed, $\Pi$ is the robot action primitive skills.

\subsection{Scene Perception Module Based on Object-centric Spatial Relation Graph}
\begin{figure}[htbp]
  \centering
  \includegraphics[width=\linewidth]{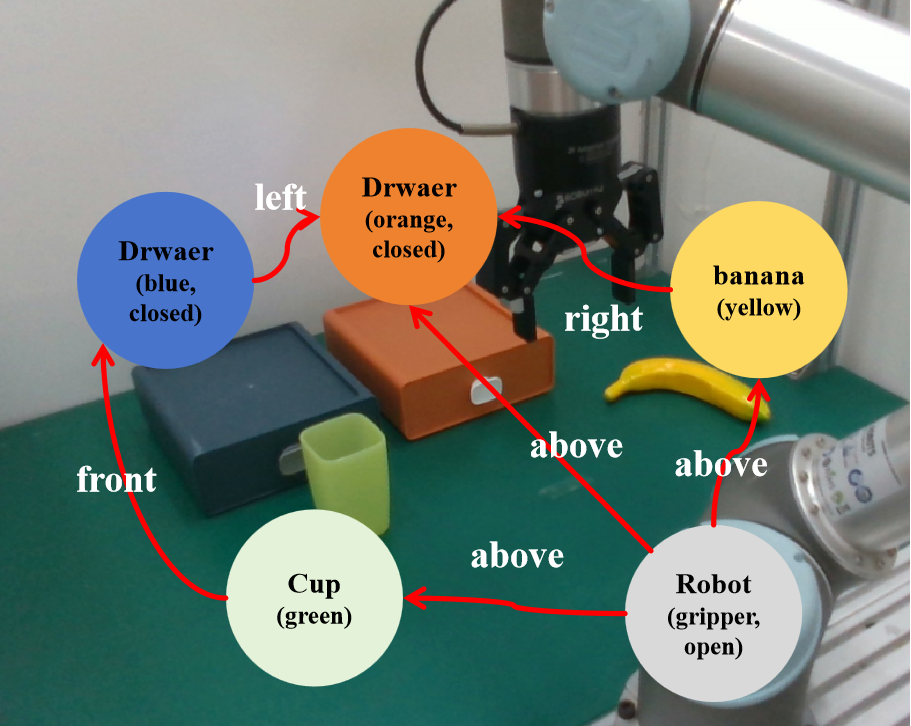}
  \caption{An example of the object-centric spatial relation graph. In the graph, the nodes represent detected objects annotated with semantic attributes, while directed edges encode their spatial relationships.}
  \label{fig:relation_graph}
\vspace{-0.5em}
\end{figure}

The core of the scene perception module is the construction of the \textbf{object-centric spatial relation graph}, which provides a structured, symbolic representation of the environment. In this graph, nodes represent detected objects annotated with semantic attributes (e.g., category, color, state such as open or closed), while directed edges encode their pairwise spatial relationships which include relative positions such as \textit{above}, \textit{below}, \textit{left}, \textit{right}, \textit{front}, and \textit{behind}. This relational representation enables the planner to reason over the scene in a language-grounded, interpretable form, facilitating goal grounding, obstacle analysis, and plan validation in dynamic environments. An illustrative example of the object-centric spatial relation graph is provided in Fig. \ref{fig:relation_graph}.

To construct this graph, we adopt a two-stage visual-linguistic pipeline. First, the input RGB image is processed using GroundedSAM2 \cite{groundedsam2} for instance segmentation with a collection of names of common objects in daily life, producing object masks and class labels. This step provides an effective visual prompt for subsequent VLM interpretation of the scene. Second, the segmented image (augmented with mask annotations), a textual summary of visual elements (e.g., object names and bounding box colors), and the task instruction are jointly fed into the VLM. The VLM is prompted to infer spatial relations and generate the structured scene representation in graph form. This pipeline leverages the VLM's capabilities in spatial perception and commonsense reasoning to infer a structured, language-based representation of the 3D spatial layout from the 2D RGB input.

\subsection{Dual-Stage Task Planner with Self-Reflection}
Many daily manipulation tasks inherently require an exploration stage to identify target objects or their states. To address this,\textbf{ the dual-stage task planner with self-reflection} employs a\textbf{ two-stage planning strategy}. Given the current relation graph $G_t$ and the high-level task goal, the planner utilizes Chain-of-Thought (CoT) \cite{CoT} prompting with an LLM to generate the plan. The process begins by determining if the current information is sufficient to achieve the goal. If critical information is missing (e.g., the location of the target object or its state), an exploration stage (\texttt{explore\_plan}) is generated. This stage defines specific exploration sub-goals, such as locating a target object, confirming its state, or identifying a specific container's contents, which are prerequisites for the main task. The completion stage (\texttt{complete\_plan}) defines the concrete sub-goal representing the actions needed to achieve the potentially abstract original goal, will be informed by the exploration results. Feedback $F_t$ from the execution validator is also input to the planner. A verification result of "yes" during the exploration stage indicates that the required missing information has been successfully acquired. The planner integrates this new knowledge into the completion stage plan, making it fully actionable, and no need for exploration plan. A "yes" result during the completion stage indicates successful achievement of the final task goal, prompting the planner to optionally perform post-task handling (e.g., scene restoration). In contrast, a "no" result at any stage will help task replann based on the failure reason, which may involve re-executing the step with adjustments, proceeding to the next step, or revising the plan logic entirely.
\begin{figure*}[thbp]
  \centering
  \includegraphics[width=\linewidth]{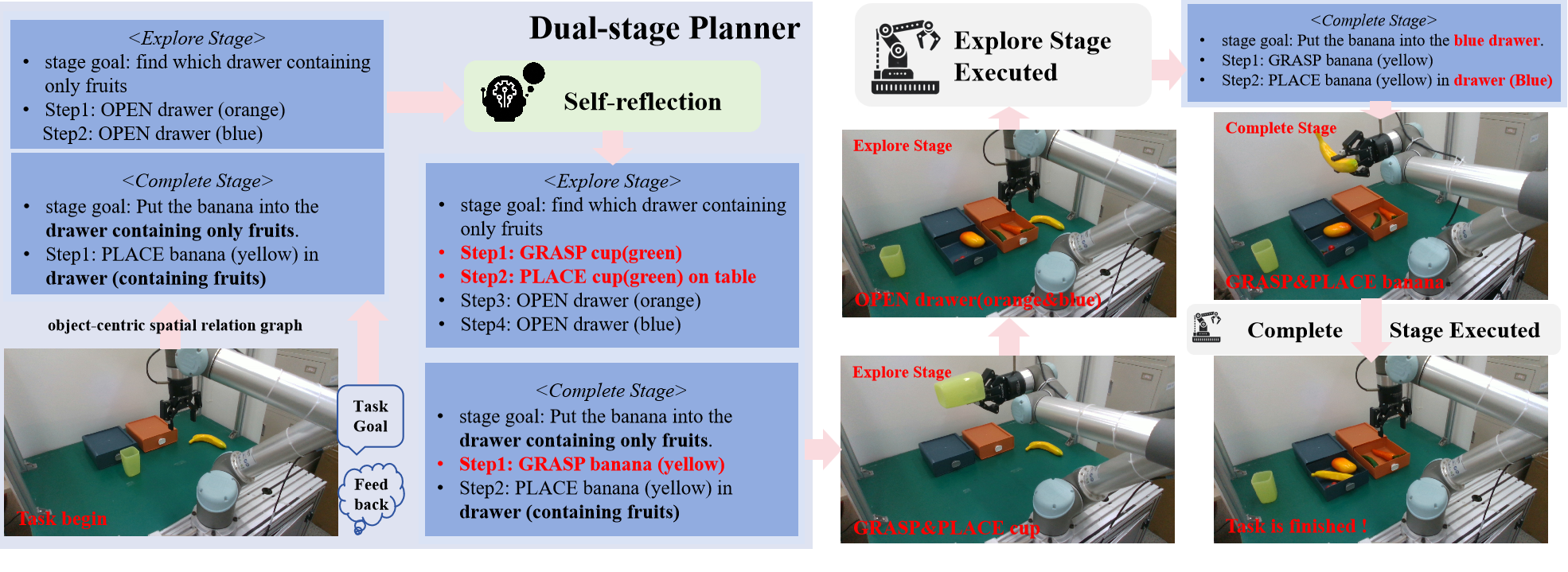}
  \caption{\textbf{An example of the dual-stage task planner with self-reflection.} The left block shows a complete thinking process of the planner. Inputs include the scene graph, the task goal $g$ (“Put the fruits into the drawer containing only fruits.”), and any available feedback. An initial plan is produced by a CoT-guided LLM that reasons over the object-centric spatial relation graph: it recognizes the banana on the table as \emph{fruit}, but the required \emph{drawer containing only fruits} remains unknown. Accordingly, the exploration sub-goal $g_E$ is to identify which drawer contains only fruits, with an exploration plan that opens both drawers. In parallel, the complete stage sub-goal $g_C$ is “Put the banana into the drawer containing only fruits”; within the corresponding complete plan, the target drawer is temporarily denoted as “containing fruits” as a placeholder that \emph{waits for exploration} to resolve. During self-reflection, two errors are corrected: (\emph{a}) using the relation graph, the cup in front of the blue drawer is flagged as an obstacle and is scheduled for removal \emph{before} opening the drawer; (\emph{b}) a missing \emph{GRASP} preceding the \emph{PLACE} of the banana is inserted. After executing the exploration stage (step-wise loops omitted), the planner updates the plan, fixes the target as the \emph{blue} drawer, and no need for a explore stage. Then the final complete stage plan is then executed (step-wise loops omitted), place the banana into the blue drawer, successfully finishing the task.}
  \label{fig:planner}
\vspace{-0.5em}
\end{figure*}

To mitigate the impact of LLM hallucinations, particularly in complex scenarios, the initial plan undergoes \textbf{a critical self-reflection step}. This mechanism rigorously checks the plan against several criteria: (a) The validity of stage goals, ensuring the exploration stage sub-goals are achievable and that achieving them enables the successful completion of the completion stage sub-goal, which in turn fulfills the original instruction; (b) The logical consistency of the action sequence, verifying prerequisites like ensuring an object is GRASPed before being PLACEd, confirming the gripper is empty before a GRASP, and checking that OPENed objects are eventually CLOSEd; (c) Obstacle handling, considering that the robotic arm operates primarily from the front; objects spatially in front of a manipulation target in the object-centric spatial relation graph are identified as potential obstacles that must be cleared (GRASPed and PLACEd elsewhere) before the target action can proceed. This self-reflection step significantly improves the correctness and robustness of the generated plan. An example is illustrated in Fig. \ref{fig:planner}.
\vspace{-0.5em}
\subsection{Execution Validator}
Delaying feedback across multiple execution steps can lead to the accumulation of small errors, ultimately resulting in major failures that are difficult to diagnose and recover from. Furthermore, interactive exploration requires timely updates in response to evolving environments. To address these challenges, the execution validator implements step-wise verification to assess the outcome of each action.

After executing a primitive action $p_t$, the validator compares the spatial relation graphs $G_t$ and $G_\mathrm{t+1}$ captured immediately before and after the step $p_t$. It then performs a semantic consistency check between the expected sub-goal $g_t$ and the observed environmental change. This process proceeds in two stages: (1) determining whether the action $p_t$ itself was successfully completed; and (2) verifying whether the corresponding sub-goal $g_t$ was achieved.
If the observed change aligns with the expected outcome, the validator outputs a verification result of "yes". Otherwise, it returns "no" along with a specific failure reason $R_t$, such as: failure of the primitive action, the need for additional steps to complete the sub-goal, logical inconsistencies in the plan, or the presence of unaddressed obstacles.

The validator's output $F_t$ is fed back to the dual-stage task planner to guide subsequent planning decisions. This \textbf{step-wise feedback} enables closed-loop adaptation and supports real-time responsiveness during task execution.

\begin{figure*}[t]
  \centering
  \begin{subfigure}[b]{0.3\textwidth}
    \includegraphics[width=\linewidth]{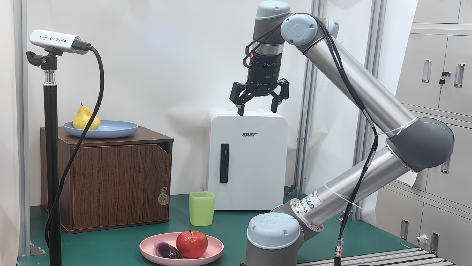}
    \caption{Overview of the system setup}
    \label{fig:a}
  \end{subfigure}
  \hfill
  \begin{subfigure}[b]{0.3\textwidth}
    \includegraphics[width=\linewidth]{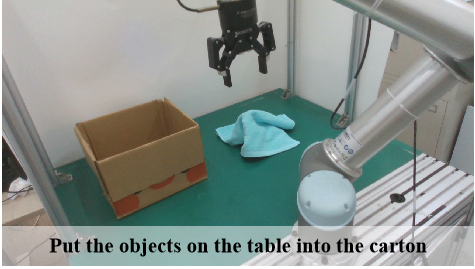}
    \caption{Task 1}
    \label{fig:b}
  \end{subfigure}
  \hfill
  \begin{subfigure}[b]{0.3\textwidth}
    \includegraphics[width=\linewidth]{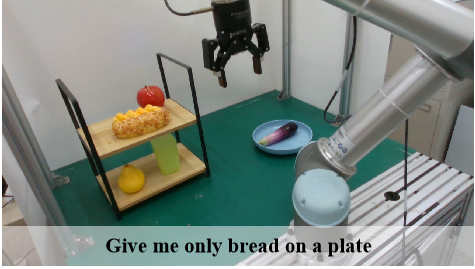}
    \caption{Task 2}
    \label{fig:c}
  \end{subfigure}

  \vspace{0.5em}

  \begin{subfigure}[b]{0.3\textwidth}
    \includegraphics[width=\linewidth]{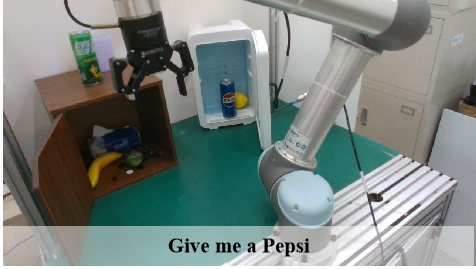}
    \caption{Task 3}
    \label{fig:d}
  \end{subfigure}
  \hfill
  \begin{subfigure}[b]{0.3\textwidth}
    \includegraphics[width=\linewidth]{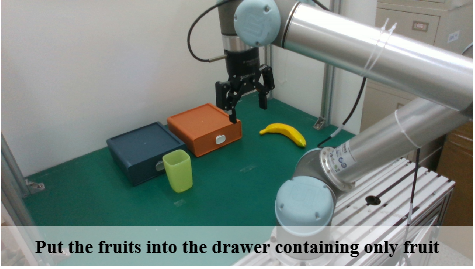}
    \caption{Task 4}
    \label{fig:e}
  \end{subfigure}
  \hfill
  \begin{subfigure}[b]{0.3\textwidth}
    \includegraphics[width=\linewidth]{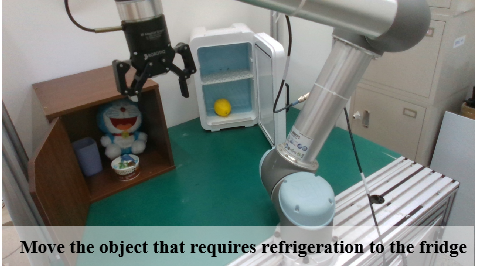}
    \caption{Task 5}
    \label{fig:f}
  \end{subfigure}

  \caption{Overview of the system and tasks setup.}
  \label{fig:all}
  \vspace{-1.5em}
\end{figure*}

\section{Experiments}
\subsection{Real-World Experimental Setup}
The experimental validation was conducted on a real robot platform. As illustrated in Fig. \ref{fig:a}, the hardware setup utilized a UR5 robotic arm equipped with a Robotiq 2F-85 gripper. Scene perception was provided by an Intel RealSense D435 RGB-D camera positioned to capture the entire operational workspace. A variety of common household objects were employed in the tasks. 

For software, GPT-4o served as the base Vision-Language Model (VLM). The LLM decomposes the goals for both stages into sequences composed exclusively of four predefined primitive robot actions:\textit{ GRASP, PLACE, OPEN, CLOSE}. These four primitive actions were implemented using established motion-primitive implementations from prior work~\cite{xu2025_gsp, xu2025align}, with scripted controllers provided for the remaining cases to maintain a consistent API. As this study focuses on high-level task planning and execution, the implementation details and reliability of these low-level primitives are not elaborated here; it is assumed that without injected noise, these primitives execute successfully.

To rigorously evaluate ExploreVLM, five distinct tasks were designed with increasing complexity:
\begin{itemize}
\item Task 1 (Basic Execution and Feedback): "Put the objects on the table into the carton." A toy hidden under a blue towel tests the system's ability to handle feedback and recover from incomplete perception. Show in Fig. \ref{fig:b}
\item Task 2 (Task Logic): "Give me only bread on a plate." Requires reasoning to remove an eggplant initially occupying the plate before placing the bread, testing task logic understanding. Show in Fig. \ref{fig:c}
\item Task 3 (Exploration and Identification): "Give me a Pepsi." Requires exploring a wooden cabinet and a refrigerator to locate the Pepsi can while ignoring distractors like a Sprite can, testing exploration and object identification. Show in Fig. \ref{fig:d}
\item Task 4 (Exploration and Commonsense): "Move the object that requires refrigeration to the fridge." Requires opening a cabinet to identify which object requires refrigeration based on commonsense, then placing it inside the fridge, testing exploration and commonsense reasoning. Show in Fig. \ref{fig:e}
\item Task 5 (Complex Exploration and Interaction): "Put the fruits into the drawer containing only fruits." Requires opening two drawers to identify the one containing fruits, removing an obstructing cup blocking the drawer, and finally placing the banana inside the right drawer, testing comprehensive capabilities including exploration, precise scene perception, obstacle handling, and commonsense. Show in Fig. \ref{fig:f}
\end{itemize}

To further assess the robustness of the closed-loop correction mechanism, a 50\% random failure rate was injected into the execution of the primitive actions (GRASP and OPEN) across all tasks. Each task was executed 10 times, and the task success rate (\%) was recorded as the primary performance metric.

\subsection{Main Results}

To evaluate the performance of the proposed ExploreVLM, we selected ReplanVLM and VILA as baseline methods. To ensure a fair comparison, all methods utilized GPT-4o as VLM. The quantitative results are presented in Table \ref{table_1}.
\begin{table}[htbp]
\caption{RESULTS OF COMPARATIVE STUDIES}
\label{table_1}
\centering
\begin{tabular}{
    >{\centering\arraybackslash}p{2cm} 
    >{\centering\arraybackslash}p{1.5cm}
    >{\centering\arraybackslash}p{1.5cm}
    >{\centering\arraybackslash}p{1.5cm}
}
\toprule
Task & ReplanVLM & VILA & OURs \\
\midrule
Task 1 & 50\% & 60\% & 100\% \\
Task 2 & 50\% & 30\% & 100\% \\
Task 3 & 10\% & 50\% & 100\% \\
Task 4 & 0\% & 10\% & 90\% \\
Task 5 & 0\% & 0\% & 80\% \\
\addlinespace[0.3em]
Average & 22\% & 30\% & 94\% \\
\bottomrule
\end{tabular}
\vspace{-1em} % 额外减少表格底部空间
\end{table}

The experimental results demonstrate that ExploreVLM achieves a 94\% average success rate, significantly outperforming both ReplanVLM (22\%) and VILA (30\%). Notably, ExploreVLM surpasses the baselines on every individual task.Analysis of the baselines' lower performance reveals several key reasons attributable to their limitations:

(a) Insufficient scene understanding led to critical oversights, such as failing to detect the eggplant obstructing the plate in Task 2, mistaking Sprite for Pepsi in Task 3, and missing the cup obstructing the drawer in Task 5. In contrast, ExploreVLM's \textbf{object-centric spatial relation graph} provided explicit and accurate spatial awareness. 

(b) Cannot handle exploratory tasks. ReplanVLM frequently entered deadlock loops when target objects have not been identified, resulting in poor performance (0-10\%) on Tasks 3-5. VILA struggled to coherently link exploration actions to execution; for instance, in Task 5, it would place the banana into the first opened drawer regardless of its contents. ExploreVLM's \textbf{dual-stage planning} with integrated real-time feedback effectively resolves this by mandating exploration before execution and enabling target confirmation during interaction. 

(c) Illogical action sequences were common in the baselines, such as attempting OPEN immediately after GRASP, not removing obstacles before operation. ExploreVLM's \textbf{self-reflection mechanism} reliably corrected such errors during planning. 

(d) Poor resilience to low-level noise was a major drawback. ReplanVLM's reliance on end-of-stage feedback allowed errors to compound. VILA often misinterpreted failed actions (e.g., an unsuccessful GRASP) as successful, leading to cascading failures (e.g., attempting to PLACE nothing) and unnecessary steps. ExploreVLM's \textbf{per-step execution verification and immediate feedback loop} proved highly effective in detecting and recovering from such noise, contributing to its superior success rate and efficiency.

\subsection{Ablation Study}

An ablation study was conducted on Task 5 (the most complex scenario) to analyze the contribution of each core module within ExploreVLM. The scene perception module (object relation graph) and the execution validator (step checker) were completely removed. The dual-stage task planner with self-reflection could not be entirely removed (as the task would be unsolvable); instead, it was replaced with a basic LLM planner that generated a standard single-stage action sequence directly from the instruction and a basic object list, bypassing the dual-stage decomposition and self-reflection. The ablation results are shown in Table \ref{table_2}. The results conclusively show that removing any of the three core modules drastically degrades performance:

\begin{table}[htbp]
\caption{RESULTS OF ABLATION STUDIES}
\label{table_2}
\setlength{\belowcaptionskip}{5pt} % 减小标题下方的间距
\centering % 代替center环境
\begin{tabular}{cc}
\toprule
Method & Success (\%)\\
\midrule
ExploreVLM & 80\%\\
ExploreVLM  w/o Object-centric Spatial Relation Graph & 30\%\\
ExploreVLM  w/o Dual-stage Task Planner with Self-reflection & 10\%\\
ExploreVLM  w/o Execution Validator & 0\%\\ 
\bottomrule
\end{tabular}
\vspace{-0.9em} % 额外减少表格底部空间
\end{table}

(a) Removing the scene perception module based on the object-centric spatial relation graph led to failures such as the inability to detect the green obstructing cup, the incapacity to confirm the contents of the correctly opened fruit drawer (preventing successful exploration), and catastrophic misidentification errors (e.g., mistaking the grasped cup for the banana). 

(b) Removing the Dual-Stage Planning with Self-Reflection resulted in the inability to determine the correct fruit drawer through exploration. The absence of self-reflection caused severe logical errors, such as sequences attempting to open a drawer while still holding the banana in the gripper and failing to clear obstacles. 

(c) Removing the Execution Verification Module eliminated the crucial per-step feedback, making it impossible to achieve a closed-loop workflow. This manifested as an inability to incorporate information discovered during exploration (e.g., which drawer contains fruit) and poor recovery from the injected primitive action noise. 

The substantial performance drop observed in each ablation condition underscores the critical and synergistic role played by all three modules – the object-centric spatial relation graph for scene perception, the dual-stage planner with self-reflection for robust high-level plan strategy, and the step execution validator for closed-loop workflow – in the overall success of the ExploreVLM framework. The success of the full system hinges on their integrated operation.

\section{CONCLUSIONS}

We presented ExploreVLM, a closed-loop robot task planning framework integrating Vision-Language Models to address the limitations of prior methods in complex interactive tasks, particularly concerning perception inadequacy, plan generation difficulty, and feedback delay. ExploreVLM leverages an object-centric spatial relation graph for precise perception, a self-reflective dual-stage (explore-then-execute) planner for handling unknowns, and a step-by-step execution feedback loop for real-time adaptation. Extensive real-robot experiments demonstrate ExploreVLM's superior performance over state-of-the-art baselines. Ablation studies rigorously validated the essential contribution of each proposed module.

Future work includes enhancing module capabilities (e.g., fine-tuning), extending ExploreVLM to multi-robot collaboration scenarios, and testing in more open-ended task environments to further advance autonomous robot intelligence in the real world. 

\addtolength{\textheight}{-12cm}   % This command serves to balance the column lengths

%%%%%%%%%%%%%%%%%%%%%%%%%%%%%%%%%%%%%%%%%%%%%%%%%%%%%%%%%%%%%%%%%%%%%%%%%%%%%%%%

%%%%%%%%%%%%%%%%%%%%%%%%%%%%%%%%%%%%%%%%%%%%%%%%%%%%%%%%%%%%%%%%%%%%%%%%%%%%%%%%

%%%%%%%%%%%%%%%%%%%%%%%%%%%%%%%%%%%%%%%%%%%%%%%%%%%%%%%%%%%%%%%%%%%%%%%%%%%%%%%%

%%%%%%%%%%%%%%%%%%%%%%%%%%%%%%%%%%%%%%%%%%%%%%%%%%%%%%%%%%%%%%%%%%%%%%%%%%%%%%%%

\end{document}